\documentclass[conference]{IEEEtran}
\IEEEoverridecommandlockouts
\usepackage{cite}
\usepackage{amsmath,amssymb,amsfonts}
\usepackage{algorithmic}
\usepackage{multirow}
\usepackage{caption}
\usepackage{subcaption}
\usepackage{graphicx}
\usepackage{textcomp}
\usepackage{xcolor}
\def\BibTeX{{\rm B\kern-.05em{\sc i\kern-.025em b}\kern-.08em
    T\kern-.1667em\lower.7ex\hbox{E}\kern-.125emX}}
\begin{document}

\title{Epidemic Modeling using Hybrid of Time-varying SIRD, Particle Swarm Optimization, and Deep Learning}

\author{\IEEEauthorblockN{Naresh Kumar}
\IEEEauthorblockA{\textit{Department of Information Technology} \\
\textit{Delhi Technological University}\\
Delhi, India \\
naresh.mehra1987@gmail.com}
\and
\IEEEauthorblockN{Seba Susan}
\IEEEauthorblockA{\textit{Department of Information Technology} \\
\textit{Delhi Technological University}\\
Delhi, India \\
seba\_406@yahoo.in}
}

\maketitle

\begin{abstract}
Epidemiological models are best suitable to model an epidemic if the spread pattern is stationary. To deal with non-stationary patterns and multiple waves of an epidemic, we develop a hybrid model encompassing epidemic modeling, particle swarm optimization, and deep learning. The model mainly caters to three objectives for better prediction: 1. Periodic estimation of the model parameters. 2. Incorporating impact of all the aspects using data fitting and parameter optimization 3. Deep learning based prediction of the model parameters. In our model, we use a system of ordinary differential equations (ODEs) for Susceptible-Infected-Recovered-Dead (SIRD) epidemic modeling, Particle Swarm Optimization (PSO) for model parameter optimization, and stacked-LSTM for forecasting the model parameters.  Initial or one time estimation of model parameters is not able to model multiple waves of an epidemic. So, we estimate the model parameters periodically (weekly). We use PSO to identify the optimum values of the model parameters. We next train the stacked-LSTM on the optimized parameters, and perform forecasting of the model parameters for upcoming four weeks. Further, we fed the LSTM forecasted parameters into the SIRD model to forecast the number of COVID-19 cases. We evaluate the model for highly affected three countries namely; the USA, India, and the UK. The proposed hybrid model is able to deal with multiple waves, and has outperformed existing methods on all the three datasets. 
\end{abstract}


\section{Introduction}
COVID-19 is one of the worst pandemics in the history. Severe Acute Respiratory Syndrome Coronavirus 2 (SARS-CoV-2) was responsible for the COVID-19 pandemic. This virus causes common cold, respiratory problems, fever, dry cough, tiredness, sore throat, body pain, head ache, in some cases diarrhea, nausea, loss of sense of taste and smell, and many more health issues which can lead to death also \cite{poletti2020probability, rabi2020sars, sarker2020self}. COVID-19 had emerged as a major threat to human race in health, social, and economical aspects of life. Many pharmaceutical and non-pharmaceutical (lockdown, social distancing, face masking, quarantine, restriction policies) methods have been adopted to fight against the pandemic. Initially, complete lockdown had been imposed to contain the spread of the virus. But imposing complete lockdown for a long time is not a good solution because it may impact emotional and economical aspect of the public. Governments should analyze the situation at ground level and implement restriction policies in the affected area only. 

Epidemiological compartmental models are commonly used for modeling the spread of an epidemic or a pandemic \cite{gordis2013epidemiology}. In these models, the population is assigned to specific labels, such as susceptible, exposed, infected, suspected, asymptomatic, hospitalized, recovered, dead etc. \cite{de2020seiard, zelenkov2023analysis}. These models use stochastic frameworks to forecast specific measures, such as the total number of infected, hospitalized, recovered, or dead people. The basic compartmental model is Susceptible-Infected-Removed (SIR) model \cite{kermack1991contributions} which is further expanded by dividing the removed compartment into recovered (R) and dead (D) compartments which is known as the SIRD model \cite{fernandez2022estimating}. It is assumed that an infected individual infects a susceptible one at a given rate $\beta$. Once infected, the individual is removed from the susceptible compartment, and enters the infected compartment. Each infected person runs through the course of the disease, and eventually is removed from the infected compartment either by recovery or death. The removed people are considered permanently immune. Most of the compartmental models consider the disease spread inside a single population and a fixed region. 

Prediction techniques are powerful tools for analyzing and planning management strategies. Many research studies based on compartmental/epidemic models have been conducted to evaluate the COVID-19 outbreak \cite{maniamfu2023lstm, karunakar2023stability}. Researchers have also enhanced epidemiological models by introducing new compartments and applying various machine learning techniques for better prediction accuracy \cite{ding2021prediction, bousquet2022deep, narayan2022using}. Shinde et al. \cite{shinde2020forecasting} summarized various forecasting techniques for COVID-19 that include stochastic theory, mathematical models, data science, and machine learning techniques. Modeling the data of COVID-19 using various statistical, prediction, and machine learning techniques can help to plan strategies to fight against a pandemic like COVID-19. A single prediction technique may not perform well in all scenarios. So, we need to develop a hybrid model to take advantage of various techniques. A hybrid model consists of multiple models or algorithms whose predictions are fused to obtain better performance.

Many mutants and variants of the SARS-Cov-2 have been reported from different countries due to which multiple waves of the COVID-19 spread was experienced by most of the countries in the world. There are multiple factors like environmental conditions (temperature, humidity, wind), mobility, lockdown, government policies, social distancing, isolation, health, age, gender, incubation period, awareness, fear, wearing of masks, hygiene, medical facilities, sanitization, testing, and vaccination which affect the spread pattern of the COVID-19. The related data can be used to find the root cause of the spread of the virus but accurate data collection is a big challenge.

In this work, we develop a hybrid model incorporating SIRD epidemic model, particle swarm optimization (PSO), and stacked Long Short-Term Memory (stacked-LSTM) neural network for COVID-19 pandemic modeling and forecasting. We estimate the SIRD model parameters weekly to accommodate the recent changes in the time series of the COVID-19 spread. The periodic estimation of the parameters is able to cater to the emergence of new waves due to new variants, or changes in the government policies. We optimize the model parameters using particle swarm optimization evolutionary algorithm. Fitting of the model parameters can be performed on available data only. So, stacked-LSTM neural network is used to identify model parameters for future prediction. Stacked-LSTM is trained on the optimized parameters of SIRD, and parameter forecasting for four weeks is performed. Further, the forecasted model parameters are fed into the SIRD model to get the forecasted values of the time series data. Thus, the main objectives of this paper are listed as follows. 
\begin{enumerate}
 \item Incorporating time-varying model parameters by estimating the parameters periodically (weekly). 
 \item Incorporating the impact of all the aspects by fitting on the actual values without collecting/inducing multi-factor data.
 \item Optimization using PSO to find out optimum values of the SIRD model parameters.
 \item Use of deep learning technique (LSTM) to forecast values of the model parameters.
 \item COVID-19 pandemic modeling using optimized SIRD model parameters for improved forecasting.
\end{enumerate}

We have adopted time series data of COVID-19 cases from the highly affected three countries namely the USA, India, and the UK. We have taken the most recent COVID-19 data containing the latest two waves from each country. We compare the proposed model with two existing methods:- standard stacked-LSTM, and hybrid of SIRD and PSO. The proposed model outperformed the two compared models on all the datasets.

Organization of the remaining paper is as follows. Related work is presented in section \ref{related_work}. Materials and proposed model are presented in section \ref{materials_methods}. Results and discussion are presented in section \ref{results_discussion}. Conclusions and future directions are presented in section \ref{conclusions}.

\section{Related Work} \label{related_work}
A number of studies based on compartmental models have been explored by the researchers to model the spread trend of the COVID-19 \cite{singh2022generalized, ning2023epi, shi2023big}. The Susceptible-Infected-Recovered (SIR) \cite{kermack1991contributions} model has been extended by a number of researchers by introducing a few new compartments related to the COVID-19 spread. Neves and Guerrero \cite{neves2020predicting} have investigated the SIR model considering either asymptomatic or few symptomatic cases of COVID-19. The authors have predicted death cases after estimating the model parameters on available data of Lombardy, Italy and Sao Paulo state in Brazil. Similarly, Postnikov \cite{postnikov2020estimation} has investigated the effectiveness of the simple SIR model for COVID-19 prediction. He has evaluated and concluded that the SIR model can provide realistic predictions of the country-level spread of COVID-19. Bastos and Cajueiro \cite{bastos2020modeling} have modeled the COVID-19 spread using two variations of the SIR model. They have included social distancing parameter in the epidemic model. Forecasting results have shown that social distancing policy is able to flatten the graph of infection cases of the COVID-19. 

Many extensions of SIRD epidemic model have been proposed to model the COVID-19 pandemic. Most of the epidemic models assume that relevant model parameters, e.g., the infection rate and the recovery rate are time-invariant, and several approaches have been proposed in the literature for tuning or estimating them \cite{sedaghat2020predicting, ghadami2022nonlinear}. Fanelli and Piazza \cite{fanelli2020analysis} have analysed the COVID-19 spread dynamics in China, France, and Italy between 22 Jan, 2020 to 15 March, 2020. A simple SIRD model was used to describe the recovery rate and death rate. The model had placed the peak in Italy around March 21, 2020. A time-dependent SIRD model was proposed in \cite{ferrari2021modeling} for provincial level COVID-19 spread prediction in Italy. Advancements in Artificial Intelligence (AI) have potential to fight against a pandemic. AI forecasting techniques can be applied for predicting the number of cases and mortality rate due to the COVID-19 pandemic. Many prediction techniques have been proposed considering various factors to improve the accuracy and matching with real scenarios \cite{kumar2020covid, kumar2021particle}. Zheng et al. \cite{zheng2020predicting} have proposed a hybrid susceptible, infected model by embedding LSTM neural network and a natural language processing module for COVID-19 prediction. The authors claimed that the proposed model can significantly reduce the prediction errors as compared to the traditional epidemic models. Researchers have investigated many machine learning and deep learning models to train prediction algorithms on COVID-19 data \cite{ayoobi2021time, xu2022forecasting, zhou2023improved, miralles2023forecasting}. The research studies are able to provide effective directions to control and manage the spread of a pandemic like COVID-19.

\section{Material and Methods}\label{materials_methods}
A compartmental model is very useful to investigate different aspects of an epidemic spread, and to show the impact of public health strategies. Basic compartmental models have the limitation of projecting a fixed pattern if the model parameters are time-invariant. To enhance the efficiency of compartmental models, and forecast realistic values, we propose a hybrid model incorporating SIRD, PSO, and stacked-LSTM model. The individual models are described in the following subsections.

\subsection{SIRD epidemic model}\label{sird_overview}
\begin{figure}[htb]
\centerline{\includegraphics[width=0.9\columnwidth]{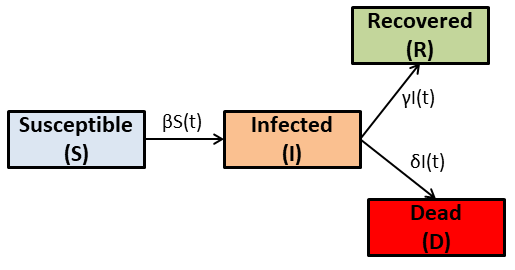}}
\caption{State diagram of SIRD model.} 
\label{fig:sird-model}
\end{figure}

We have adopted the SIRD model for COVID-19 pandemic spread modeling. The SIRD model divides the population into four compartments \textit{viz.} susceptible (S), infected (I), recovered (R), and dead (D) individuals. Each individual in the population has equal probability of being infected. State transition diagram of the model is shown in Fig. \ref{fig:sird-model}. 

Ordinary differential equations (ODEs) for the SIRD model are as follows.
\begin{equation}
\frac{dS(t)}{dt} = -\beta\frac{S(t)I(t)}{N}
\end{equation}

\begin{equation}
\frac{dI(t)}{dt} = \beta\frac{S(t)I(t)}{N} - \gamma I(t) - \delta I(t)
\end{equation}

\begin{equation}
\frac{dR(t)}{dt} = \gamma I(t)
\end{equation}

\begin{equation}
\frac{dD(t)}{dt} = \delta I(t)
\end{equation}   

Where $S(t), I(t), R(t)$, and $D(t)$ are susceptible, infected, recovered, and dead individuals at time $t$, $\beta$ is infection rate, $\gamma$ is recovery rate, and $\delta$ is fatality rate. The model is a positive system in which all variables have positive values for time $t > 0$, when initialized with no-negative values at time $t = 0$. We have omitted human births and natural deaths in the model \cite{lopez2021modified}. Sum of all the states is constant and equal to the total population $N$.

\begin{equation}
N = S(t) + I(t) + R(t) + D(t)
\end{equation}

In our approach, we have utilized the SIRD model for data fitting and data forecasting. The model parameters ($\beta$, $\gamma$, $\delta$) are optimized using PSO algorithm by fitting SIRD on real data. Description of PSO is given in the next subsection.

\subsection{Particle Swarm Optimization (PSO)} \label{pso_overview}
PSO \cite{kennedy1995particle} is an evolutionary algorithm which consists of a swarm of particles. Each particle of the swarm represents a possible solution of an optimization problem. The algorithm is able to search a nearly optimal solution of an optimization problem with a minimum computation, and without getting trapped into a local minimum solution. The particles swim through the n-dimensional search space of an optimization problem in search of an optimal solution. 

Lets a swarm consists of $P$ number of particles where position vector of $i_{th}(i=1,2, …, P)$  particle is denoted as $X_{i} = [x_{i}^{1}, , x_{i}^{2}, ..., x_{i}^{n}]$, and velocity vector as $V_{i} = [v_{i}^{1}, v_{i}^{2}, ..., v_{i}^{n}]$. Initially, random values of position vector are generated within the search space. To find the optimal solution, each particle swims through the search space and keeps the personal best position which has been reached so far. Best position among all the particles is recorded as global best. Velocity and position of a particle are updated in each iteration till maximum iteration is reached or pre-defined condition is met. Velocity and position of a particle, and inertia weight are updated using the following equations.

\begin{equation} \label{pso_intertia_weight_eqn}
 w^{t} = w_{max} - \frac{t*(w_{max} - w_{min})}{t_{max}} 
\end{equation}

\begin{multline} \label{pso_velocity_eqn}
 V_{i}^{t+1} = w^{t} * V_{i}^{t} + C1*Rand()*(P_{best\_i} - X_{i}^{t})  \\ + C2*Rand()*(P_{gbest} - X_{i}^{t})
\end{multline}

\begin{equation} \label{pso_position_eqn}
 X_{i}^{t+1} = X_{i}^{t} + V_{i}^{t+1}
\end{equation}

where $w_{min}$ and $w_{max}$ are the pre-defined inertia weights. $w^{t}$ is the inertia weight in the $t_{th}$ iteration. $V_{i}^{t}$ and $X_{i}^{t}$ are the velocity vector and position vector of the $i_{th}$ particle, respectively. $P_{best\_i}$ is the personal best of $i_{th}$ particle and $P_{gbest}$ is the global best among all the particles. Velocity of a particle is restricted to the pre-defined range $[-Vmax, Vmax]$. $Rand()$ is a random function to generate a value in the range $[0,1]$ under uniform distribution. $C1$ and $C2$ are the acceleration coefficients. $t_{max}$ is the maximum iteration count. We have used PSO algorithm to identify optimum values of SIRD model parameters which are best fitting of the COVID-19 pandemic evolution. 

\subsection{Long Shot-Term Memory (LSTM)}
Long Short-Term Memory (LSTM) neural network was proposed by Hochreiter and Schmidhuber \cite{hochreiter1997long}. The main objective of LSTM is to solve the vanishing gradient problem associated with the traditional Recurrent Neural Network (RNN). LSTM has four main components; a memory cell and three gates namely input gate, output gate, and forget gate. 
The memory cell holds information over the period of time. The input gate manages the information to be passed to the memory cell. The forget gate removes the irrelevant information from the memory cell. The output gate decides the output of the memory cell. Thus, LSTM remembers the relevant information and passes the selective information through the network. Therefore, LSTM is capable to learn long-term dependencies. Variants of LSTM such as stacked-LSTM, bidirectional-LSTM, convolutional-LSTM, and attention-LSTM have been proposed to solve the more complex problems. LSTM is widely used for time series forecasting, language translation, and speech recognition \cite{gautam2022transfer}. We have used LSTM to perform upcoming 28 days (4 weeks) forecasting of the SIRD model parameters after optimization using PSO algorithm. 

\subsection{Proposed Model}
\begin{figure}[htb]
\centerline{\includegraphics[width=0.6\columnwidth]{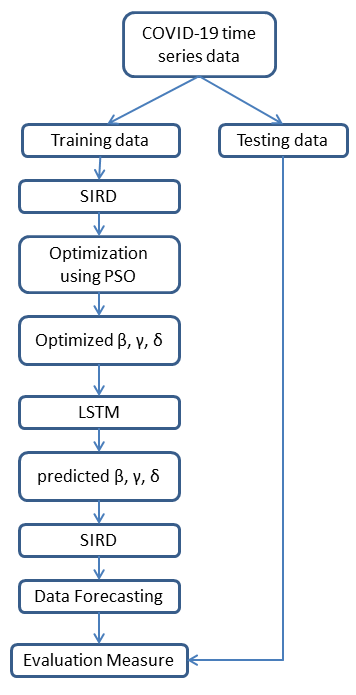}}
\caption{Steps of the proposed hybrid model.} 
\label{fig:proposed-model}
\end{figure}

We propose a hybrid model using SIRD, PSO, and stacked-LSTM to improve the forecasting accuracy. SIRD model parameters are optimized using PSO, and fitted on real values of infected, recovered, and fatality cases weekly. We initialize the model states $S(0), I(0), R(0), D(0)$ using real values at $t=0$, and $S(t), I(t), R(t), D(t)$ using values at time  $t>0$.  The weekly optimized SIRD model parameters are fed into the stacked-LSTM neural network to train and forecast the parameters for upcoming four weeks. The forecasted parameters are used in the SIRD model to get the evolution trend of the cases for next 28 days. The proposed hybrid model is shown in Fig \ref{fig:proposed-model}. Steps of the hybrid model are explained below.
\begin{enumerate}
 \item Data pre-processing:  We have used day-level cumulative cases of COVID-19. Time series data of COVID-19 is very much fluctuating in nature. So, we have performed weekly averaging for smoothening of the data. 
 \item Train-test splitting: We split the pre-processed data into train and test samples. We kept last 28 days (four weeks) samples for testing. 
 \item SIRD modeling: We have used SIRD model for the COVID-19 pandemic evolution. The model is explained in subsection \ref{sird_overview}. 
 \item Parameter optimization: We estimate the SIRD model parameters periodically. We have used PSO algorithm to optimize the SIRD model parameters ($\beta$, $\gamma$, $\delta$). Initially, the values of infection rate ($\beta$), recovery rate ($\gamma$), and fatality rate ($\delta$) are randomly generated between 0 and 1. The parameter values are updated iteratively using the equations defined in subsection \ref{pso_overview}, and fitted on real values of COVID-19 cases using SIRD model. Data fitting of $S(t), I(t), R(t), D(t)$ values are evaluated with respect to Mean Squared Error (MSE). Optimum values of the parameters ($\beta, \gamma, \delta$) are returned by the PSO. We set the periodicity of the parameters estimation as weekly in our model.
 \item Parameter forecasting: We have used stacked-LSTM to forecast the value of SIRD model parameters. The weekly optimized parameters ($\beta, \gamma, \delta$) are used to train the LSTM model, and perform forecasting for four weeks. The configuration used for LSTM is discussed in section \ref{results_discussion}.
 \item Pandemic evolution: In this step, SIRD model is utilized to identify the final crisp values of the data. Stacked-LSTM forecasted parameters are fed into the SIRD model. Four weeks forecasted values of $\beta$, $\gamma$, and $\delta$ are utilized in SIRD model to show evolution of the COVID-19 pandemic during the upcoming 28 days. 
\end{enumerate}

\subsection{Datasets}
We have used COVID-19 time series data of three countries namely the USA, India, and the UK in our experiments. The data is publicly available at github repository which is updated on a regular basis \cite{ourworldindata}. We have used the dataset of each countries  including the last two major waves of the COVID-19 spread. The detail of the timeline for each country is given in Table \ref{table:datasets}. 

\begin{table}[htbp]
\caption{COVID-19 time series of adopted countries}
\label{table:datasets}
\begin{center}
\renewcommand{\arraystretch}{1.2}
\begin{tabular}{l l l l}
\hline
Dataset & Start &	End & Population \\
\hline
USA	& July 1, 2021	& April 15, 2022 &	330 Million  \\
India &	January 1, 2021 &	March 30, 2022	& 1100 Million  \\
UK & June 1, 2021 &	April 30, 2022 & 60 Million         \\
\hline
\end{tabular}
 \end{center}
\end{table}

\subsection{Performance Measures}
To evaluate the prediction models, we use the Root Mean Square Error (RMSE) statistical measure.

\begin{equation}
RMSE = \sqrt{\frac{1}{M} \sum_{i=1}^{M} \left( y_{i} - y\hat{}_{i} \right)^{2} }
\end{equation}

where $y_{i}$ denotes actual value and $y\hat{}_{i}$ denotes predicted value for the $i_{th}$ instance. $M$ is the total number of testing samples.

\section{Experimental Setup and Results}\label{results_discussion}
The proposed hybrid model is implemented in Python 3.9. We have executed our experiments in the system having 8 GB RAM, 4GB NVIDIA GTX-1650 GPU, and Intel Core i5 processor clocked at 2.40 GHz. The forecasting results of the proposed model are compared with the standard stacked-LSTM, and hybrid of SIRD and PSO. The detail of the model configurations, results and discussion are presented in the following subsections.

\subsection{Experimental setup}
Our model is hybrid of SIRD, PSO, and LSTM. We set the PSO parameters as shown in Table \ref{table:pso-param}. We kept 200 as the maximum number of iterations for each experiment in PSO because the algorithm has shown optimum convergence in around 200 iterations. . 

\begin{table}[htbp]
\caption{PSO parameters set in the experiments}
\label{table:pso-param}
\begin{center}
\renewcommand{\arraystretch}{1.2}
\begin{tabular}{l l}
\hline
PSO Parameter & Value \\
\hline
Number of particles &	30    \\
Maximum number of iterations & 	200 \\
Inertia weight  [$w_{min},  w_{max}$] &	[0.4, 0.9] \\
C1 = C2 &	2  \\
\hline
\end{tabular}
 \end{center}
\end{table}

We have used stacked-LSTM for forecasting of SIRD model parameters. Configured parameters for LSTM are shown in Table \ref{table:dnn-param}. In stacked-LSTM model, two layers of LSTM is used; first layer with 20 hidden units, and second layer with 30 hidden units. 

\begin{table}[htbp]
\caption{LSTM parameters}
\label{table:dnn-param}
 \begin{center}
\renewcommand{\arraystretch}{1.2}
\begin{tabular}{l l}
\hline
Parameter & Value \\
\hline
Activation Function &	linear    \\
Optimizer & Adam   \\
Learning rate & 0.005 \\
Loss function  & Huber \\
Evaluation metrics & MSE \\
Epochs & 500   \\
\hline
\end{tabular}
\end{center}
\end{table}

\begin{figure}
\begin{subfigure}{.5\textwidth}
  \centering
  \includegraphics[height=0.6\columnwidth, width=.80\linewidth]{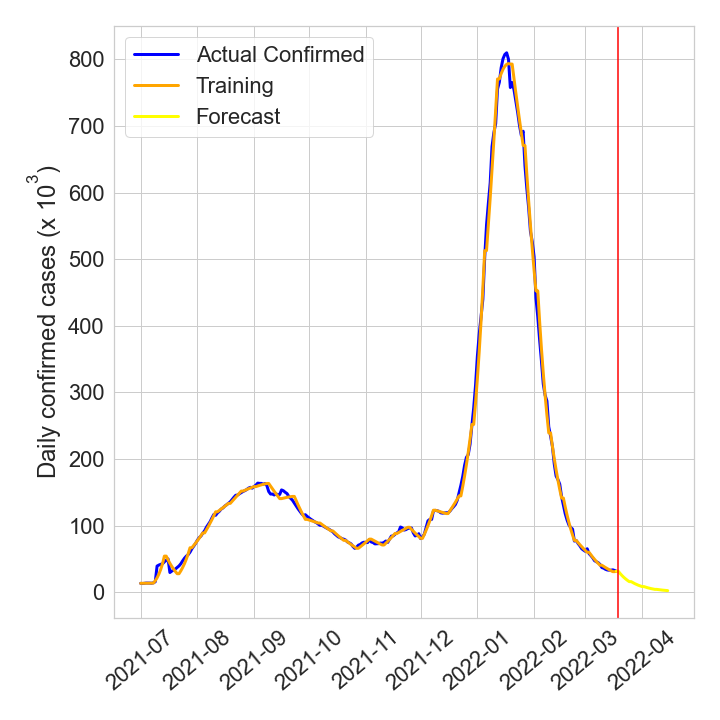}  
  \caption{USA forecast using proposed model}
  \label{fig:usa-forecast}
\end{subfigure}
\hfill
\begin{subfigure}{.5\textwidth}
  \centering
  \includegraphics[height=0.6\columnwidth, width=.80\linewidth]{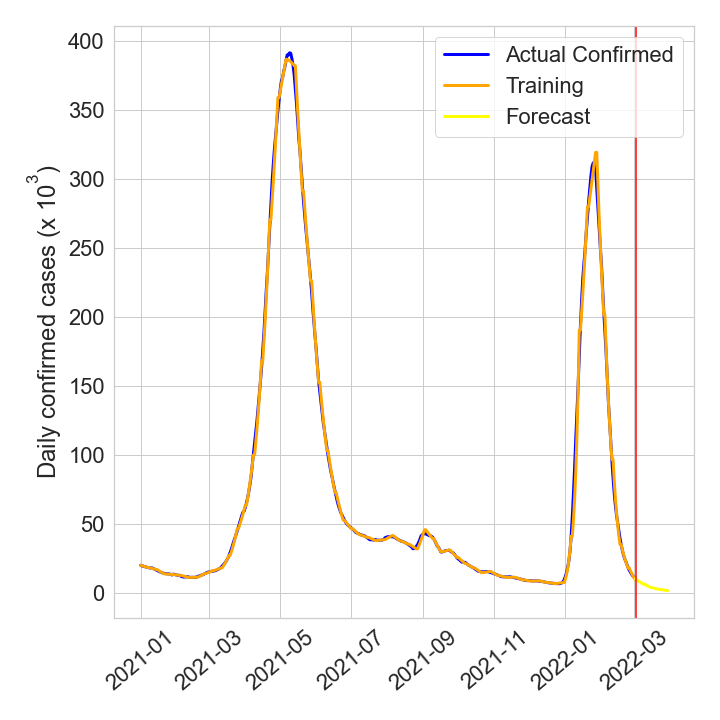}  
  \caption{India forecast using proposed model}
  \label{fig:india-forecast}
\end{subfigure}
\hfill
\begin{subfigure}{.5\textwidth}
  \centering
  \includegraphics[height=0.6\columnwidth, width=.80\linewidth]{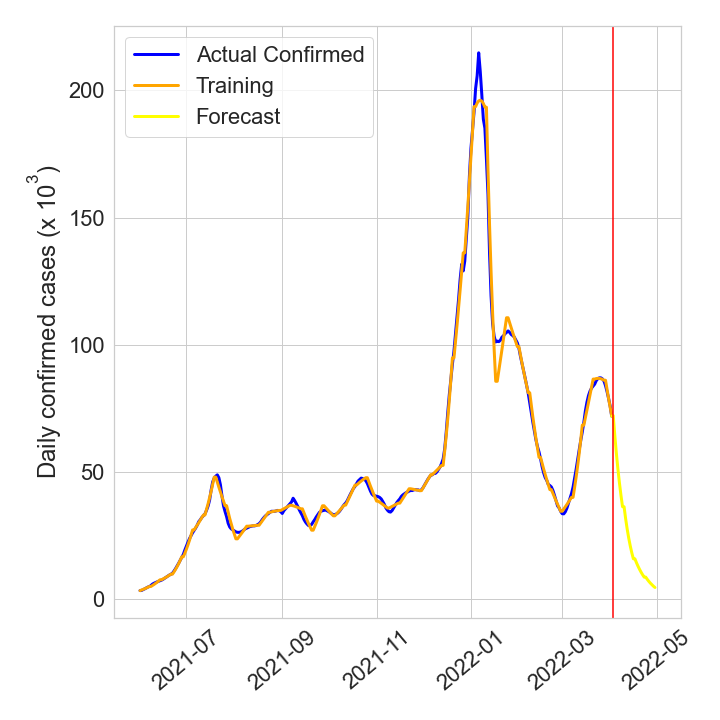}  
  \caption{UK forecast using proposed model}
  \label{fig:uk-forecast}
\end{subfigure}

\caption{Forecasting of infected cases using experimented models and comparison with test data of the adopted country.}
\label{fig:complete-forecast}
\end{figure}

\begin{figure*}
\begin{subfigure}{.3\textwidth}
  \centering
  \includegraphics[width=.95\linewidth]{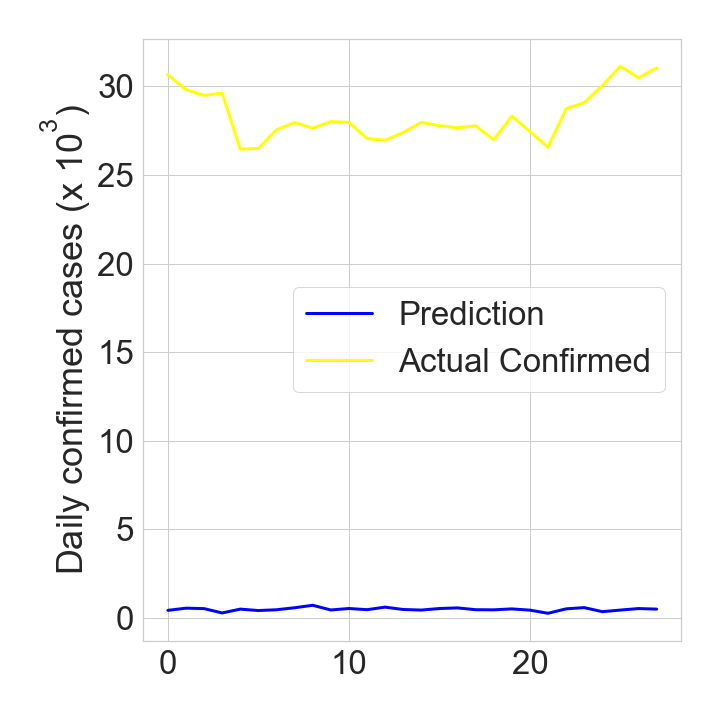}  
  \caption{USA forecast using LSTM}
  \label{fig:usa-sub-first}
\end{subfigure}
\hfill
\begin{subfigure}{.3\textwidth}
  \centering
  \includegraphics[width=.95\linewidth]{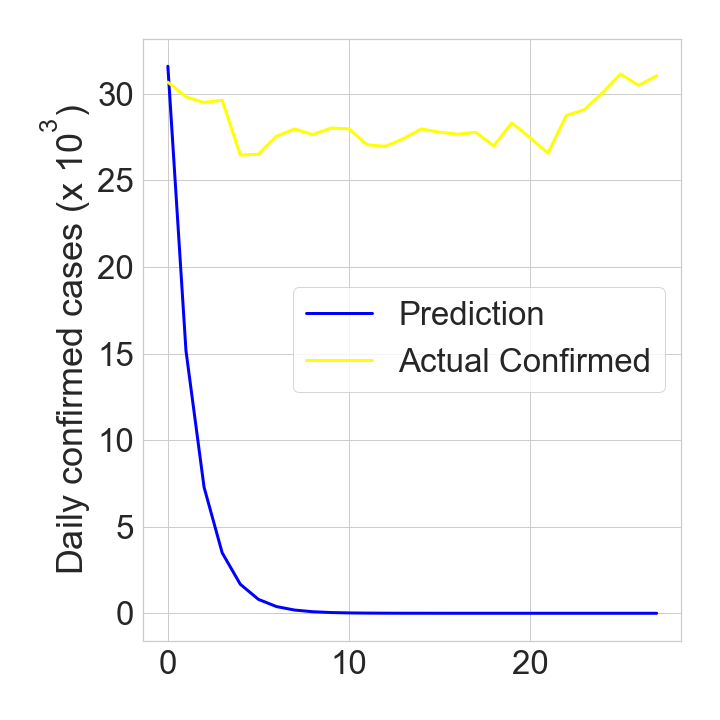}  
  \caption{USA forecast using SIRD+PSO}
  \label{fig:usa-sub-second}
\end{subfigure}
\hfill
\begin{subfigure}{.3\textwidth}
  \centering
  \includegraphics[width=.95\linewidth]{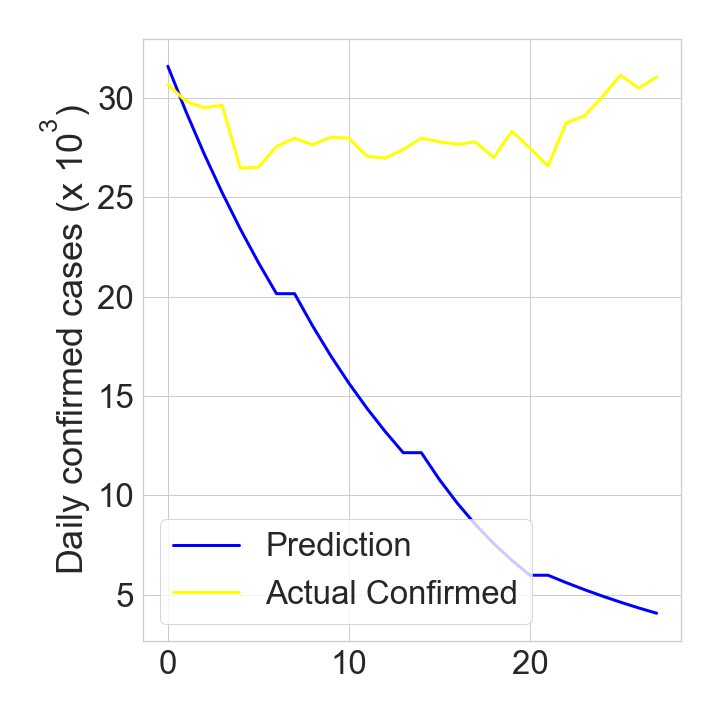}  
  \caption{USA forecast using SIRD+PSO+LSTM}
  \label{fig:usa-sub-third}
\end{subfigure}

\begin{subfigure}{.3\textwidth}
  \centering
  \includegraphics[width=.8\linewidth]{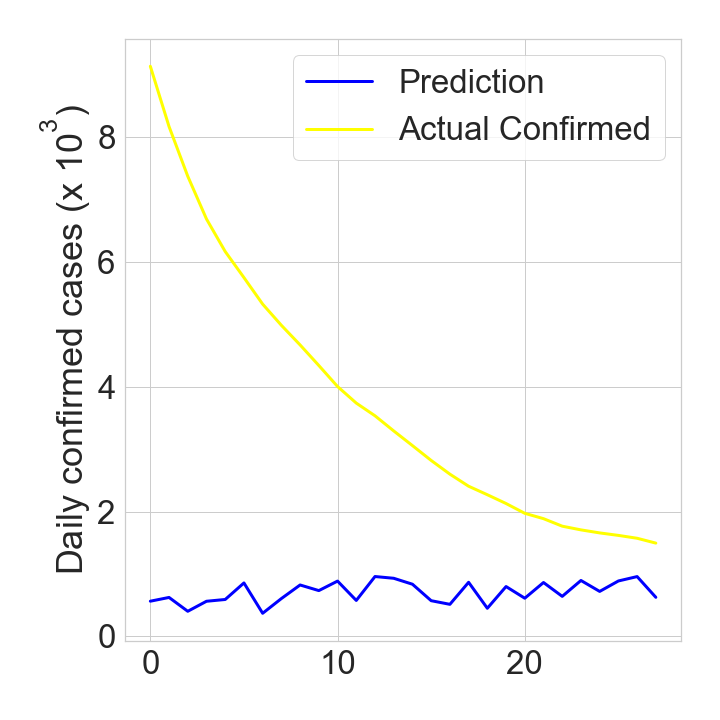}  
  \caption{India forecast using LSTM}
  \label{fig:india-sub-first}
\end{subfigure}
\hfill
\begin{subfigure}{.3\textwidth}
  \centering
  \includegraphics[width=.8\linewidth]{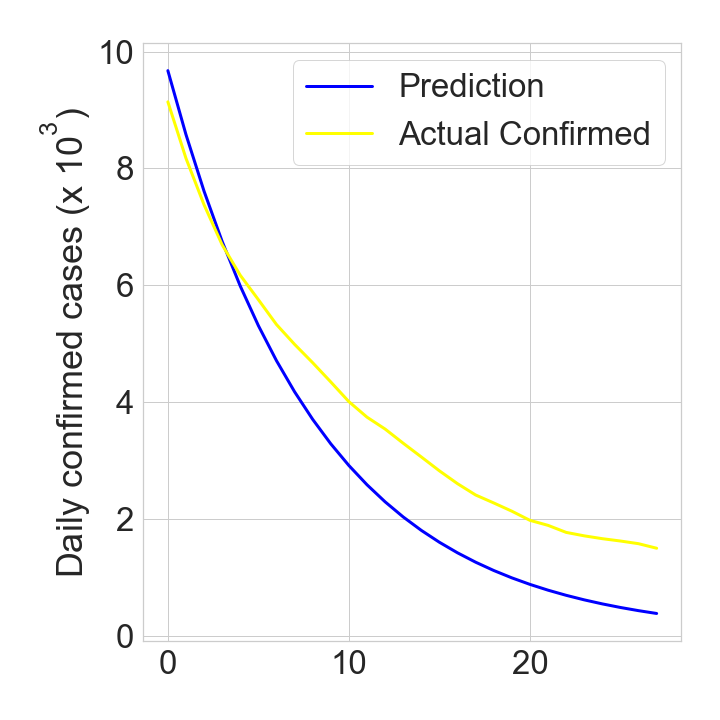}  
  \caption{India forecast using SIRD+PSO}
  \label{fig:india-sub-second}
\end{subfigure}
\hfill
\begin{subfigure}{.3\textwidth}
  \centering
  \includegraphics[width=.8\linewidth]{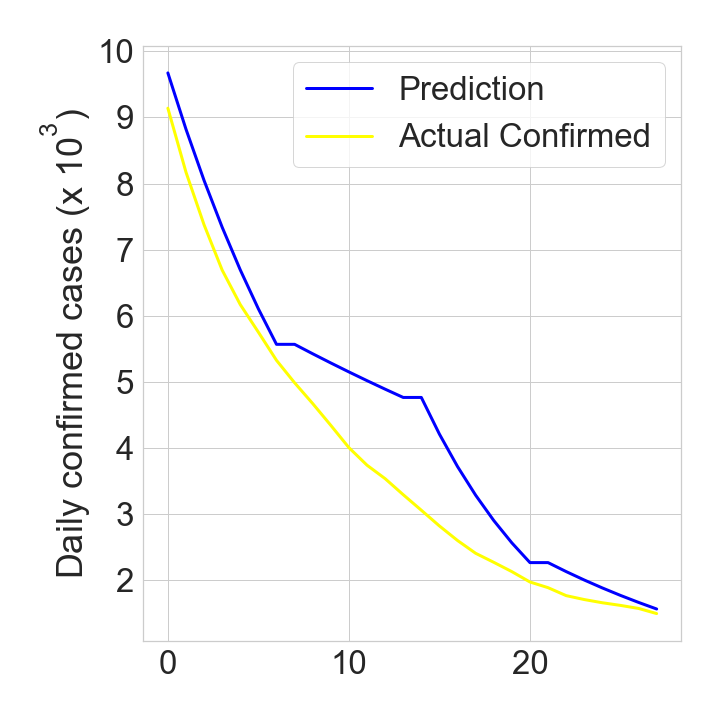}  
  \caption{India forecast using SIRD+PSO+LSTM}
  \label{fig:india-sub-third}
\end{subfigure}

\begin{subfigure}{.3\textwidth}
  \centering
  \includegraphics[width=.8\linewidth]{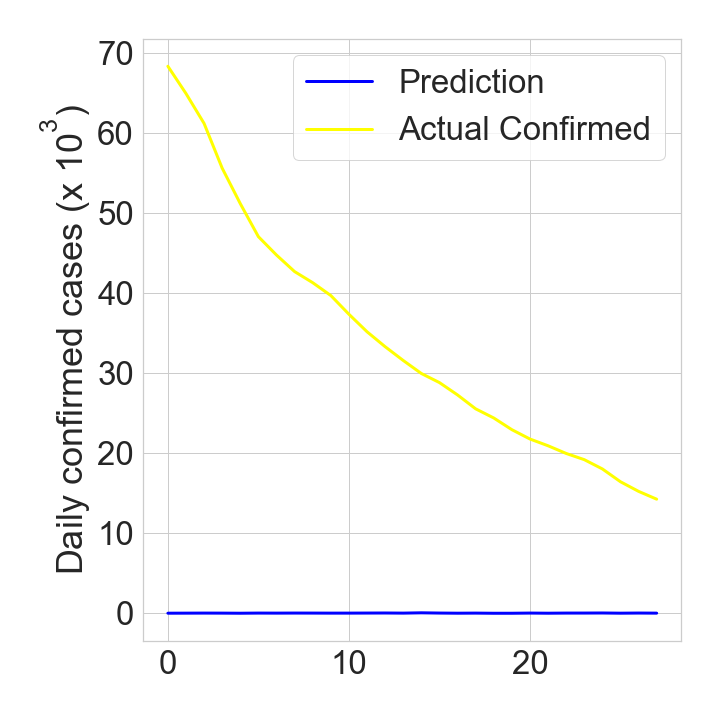}  
  \caption{UK forecast using LSTM}
  \label{fig:uk-sub-first}
\end{subfigure}
\hfill
\begin{subfigure}{.3\textwidth}
  \centering
  \includegraphics[width=.8\linewidth]{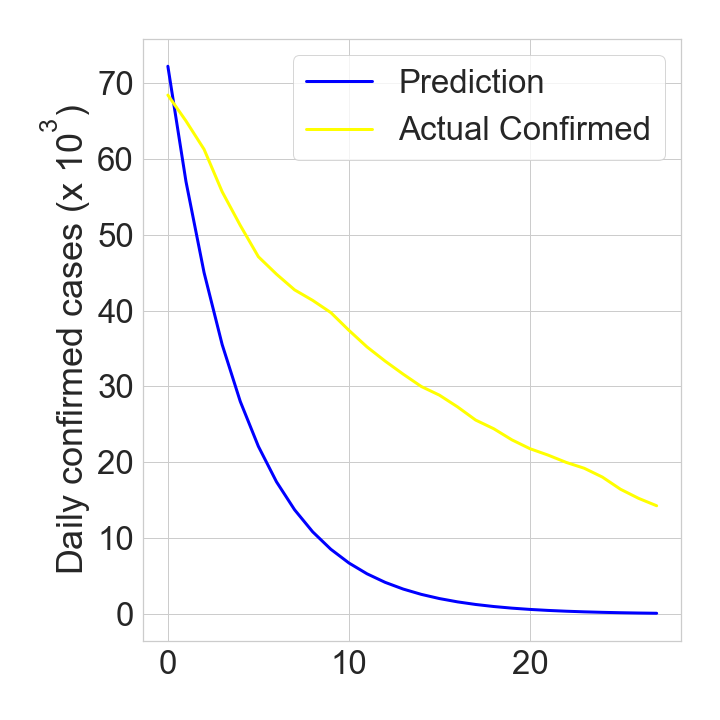}  
  \caption{UK forecast using SIRD+PSO}
  \label{fig:uk-sub-second}
\end{subfigure}
\hfill
\begin{subfigure}{.3\textwidth}
  \centering
  \includegraphics[width=.8\linewidth]{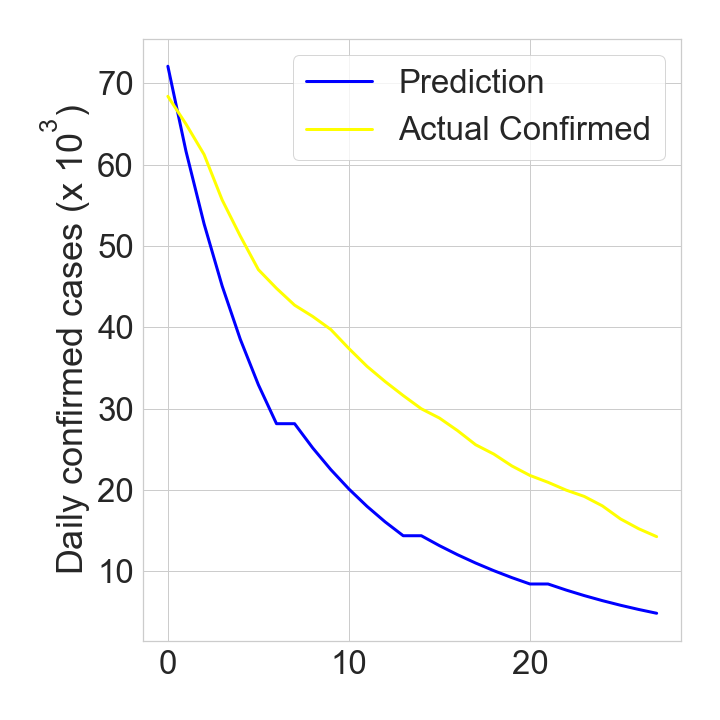}  
  \caption{UK forecast using SIRD+PSO+LSTM}
  \label{fig:uk-sub-third}
\end{subfigure}

\caption{Forecasting of infected cases using experimented models and comparison with test data of the adopted country.}
\label{fig:forecasting-trends}
\end{figure*}

\subsection{Forecasting results}
We have evaluated all the developed forecasting models for infected cases of COVID-19 in the adopted countries. We have divided the results in three ways, first we show training and forecasting results of the proposed model in graphical form, then forecasting accuracy results of all the compared models in tabular form, and graphic form. Training and forecasting results of the proposed model for all the three countries are shown in Fig \ref{fig:complete-forecast}. A vertical line is used to separate the training and forecasting results. Forecasting results are for upcoming 4 weeks.  

\begin{table}[htbp]
\caption{Forecasting accuracy (RMSE) for infected cases}
\label{table:forecast-accuracy}
 \begin{center}
\renewcommand{\arraystretch}{1.2}
\begin{tabular}{l l l l}
\hline
Model & US & India & UK\\
\hline
Stacked-LSTM	& 27836	& 3814 &	37479 \\
SIRD + PSO	& 25816 &	2399 &	26582  \\
SIRD + PSO + LSTM	& 23528	& 1887	& 24629 \\

\hline
\end{tabular}
\end{center}
\end{table}

The comparative results for all the models are shown in Table \ref{table:forecast-accuracy}. We can see that hybrid of SIRD, PSO, and stacked-LSTM model outperformed all the other compared models on COVID-19 datasets. Combination of SIRD and PSO also better performed as compared to stacked- LSTM model. 
Comparative analysis of the experimented models is given in Fig. \ref{fig:forecasting-trends} for each country. The figure shows infected cases forecasting trend for each country for upcoming four weeks. We can observe the accuracy of the models for each country from the figure that the best performing model is the proposed hybrid model. 

\section{Conclusion and Future work} \label{conclusions}
Many epidemics have been emerged over the period of time. A recent pandemic is COVID-19 which has taken the lives of millions of people. The SIRD epidemiological model alone is not able to depict the time-varying evolution trend of COVID-19 pandemic due to the dynamically evolving behavior of the pandemic. So, in this work, we have proposed a hybrid model to deal with the dynamic nature of the COVID-19 pandemic by combining the advantages of an evolutionary optimization algorithm such as PSO, and deep learning method such as stacked-LSTM neural network. Therefore, our proposed hybrid model consists of SIRD, PSO, and stacked-LSTM. We have evaluated the model on COVID-19 spread pattern. The proposed model outperformed all the other compared models on all the adopted COVID-19 datasets namely, the USA, India, and the UK. The model can be used to analyse and predict the epidemic spread pattern which can be used to guide governments to act in advance. The proposed approach can be experimented with other deep learning techniques and epidemic datasets.

\bibliographystyle{plain}
\bibliography{epidemic-modeling}

\end{document}